\documentclass[runningheads,a4paper]{llncs}

\usepackage{amssymb}
\usepackage{graphicx}
\usepackage{graphicx}
\usepackage{amsmath}
\usepackage{amssymb}
\usepackage{color}
\usepackage{algorithmic}
\usepackage{algorithm}
\usepackage{subfigure}
\usepackage{array}
\usepackage{multirow,tabularx}
\usepackage{multicol}
\usepackage{pdfsync}

\usepackage{url}
\urldef{\mailsa}\path|{alfred.hofmann, ursula.barth, ingrid.haas, frank.holzwarth,|
\urldef{\mailsb}\path|anna.kramer, leonie.kunz, christine.reiss, nicole.sator,|
\urldef{\mailsc}\path|erika.siebert-cole, peter.strasser, lncs}@springer.com|
\newcommand{\keywords}[1]{\par\addvspace\baselineskip
\noindent\keywordname\enspace\ignorespaces#1}

\newcommand{\bfb}{{\textbf{b}}}

\newcommand{\bfx}{{\textbf{x}}}

\newcommand{\bfw}{{\textbf{w}}}
\newcommand{\bfy}{{\textbf{y}}}

\newcommand{\bft}{{\textbf{t}}}

\newcommand{\bfphi}{{\boldsymbol{\phi}}}

\begin{document}

\mainmatter  

\title{A novel image tag completion method based on convolutional neural transformation}

\titlerunning{A novel image tag completion method based on convolutional neural representation}

\author{Yanyan Geng$^1$ \and Guohui Zhang$^2$ \and Weizhi Li$^3$ \and Yi Gu$^4$ \and Ru-Ze Liang$^5$ \and Gaoyuan Liang$^6$ \and Jingbin Wang$^7$ \and  Yanbin Wu$^8$  \and Nitin Patil$^9$\and Jing-Yan Wang$^{1}$
\footnote{The study was supported by Provincial Key Laboratory for Computer Information Processing Technology, Soochow University, China (Grant No. KJS1324).}
}
\authorrunning{Geng, Y, etc.}

\institute{$^1$ Provincial Key Laboratory for Computer Information Processing Technology,
Soochow University, Suzhou 215006, China\\
Email: yanyangeng@outlook.com\\
$^2$ Huawei Technologies Co., Ltd., Shanghai, China\\ 
$^3$ Suning Commerce R\&D Center USA, Inc., Palo Alto, CA 94304, United States\\
$^4$ Analytics \& Research, Travelers, Hartford, CT 06183, United States\\
$^5$ King Abdullah University of Science and Technology, Saudi Arabia\\
$^6$ Jiangsu University of Technology, Jiangsu 213001, China\\
$^7$ Information Technology Service Center,
Intermediate People's Court of Linyi City, Linyi, China\\
Email: jingbinwang1@outlook.com\\
$^8$ Hebei University of Economics and Business,
Shijiazhuang 050061, China\\
$^9$ Savitribai Phule Pune University, Pune, Maharashtra 411007, India
}

\toctitle{Lecture Notes in Computer Science}
\tocauthor{Authors' Instructions}
\maketitle

\begin{abstract}
In the problems of image retrieval and annotation, complete textual tag lists of images play critical roles. However, in real-world applications, the image tags are usually incomplete, thus it is important to learn the complete tags for images. In this paper, we study the problem of image tag complete and proposed a novel method for this problem based on a popular image representation method, convolutional neural network (CNN). The method estimates the complete tags from the convolutional filtering outputs of images based on a linear predictor. The CNN parameters, linear predictor, and the complete tags are learned jointly by our method. We build a minimization problem to encourage the consistency between the complete tags and the available incomplete tags, reduce the estimation error, and reduce the model complexity. An iterative algorithm is developed to solve the minimization problem. Experiments over benchmark image data sets show its effectiveness.
\end{abstract}

\section{Introduction}

Image tagging problem is defined as assigning a set of textual tags to an targeted set of images, and it has becoming more and more important for both image retrieval and annotation applications \cite{Fu20161985}. In the ideal case, the tags of an image should be accurate and complete. However, in the real-world applications, the tags of images are usually incomplete, and it is necessary to complete the tags of images. The problem of completing the tags of images are called image tag completion \cite{Yang20162407}. To solve this problem, many approaches have been proposed \cite{Lin20131618,Wu2013716,Feng2014424,Lin201442,Xia2015500,Li2016425}, but the performance of these works are not satisfying yet. Meanwhile, convolutional neural network (CNN) has been shown to be an effective tool to represent images \cite{geng2016learning,Lopes2017610,Ma2017221,Yang2017421}. However, surprisingly, CNN has not been applied to the problem of image tag completion.

In our paper, we propose a novel image tag completion method based on the convolutional representation of the images and the linear prediction of the tag assignment vectors. We first use a CNN model to represent the images to the convolutional vectors, and then apply a linear predictive model over the convolutional representations to obtain the complete tag assignment vectors of the images. The possible effect of using convolutional transformation lies on the potential of finding effective visual features for the task of tag prediction. To learn the parameters of the CNN and the linear predictive model, we impose the learned tag assignment vectors to be consistent to the available elements of the incomplete tag vectors, minimize the prediction errors of the linear predicative model over the image set. We also minimize the distance between tag assignment vectors images which have large convolutional similarities. Finally, we apply the sparsity penalty to the tag assignment vectors. To solve argued minimization problem, we use gradient descent method. The experimental results over some benchmark data sets show that the proposed convolutional representation-based tag completion method outperforms the state-of-the-art tag completion methods.

\section{Proposed method}
\label{sec:method}

We suppose we have a set of images, denoted as $\mathcal{I} = \{I_1,\cdots,I_n\}$, and a set of candidate tags of these images, $\mathcal{T}=\{T_1,\cdots, T_m\}$. To denote the assigning relation between the images and the tags, we define a matrix of assignment, $T=[t_{ji}]\in \{1,0\}^{m\times n}$, and its $(j,i)$-th entity $t_{ji}$ is set to $1$ if $T_j$ is assigned to $I_i$, and $0$ otherwise. $T$ is the output of the model, and in our learning process, it has been relaxed to realtime value matrix $T\in \mathbb{R}^{m\times n}$. We have a existing assignment matrix $\widehat{T} = [\widehat{t}_{ji}]\in \{1,0\}^{m\times n}$, and its entities are of the same meaning as $T$, but it is incomplete. We further define a binary matrix $\Phi = [\phi_{ji}] \in \{1,0\}^{m\times n}$, where its $(j,i)$-th entity is defined to indicate if $\widehat{t}_{ji}$ is missing, $\phi_{ji}=
0, if~\widehat{t}_{ji}~is~missing, ~and~1, otherwise.$ The problem of image tagging is transformed to the learning of a complete assignment matrix $T$ from $\mathcal{I}$, $\widehat{T}$, and $\Phi$.

To complete the tags of an image, $I$, we propose to learn its convolutional representation and the complete tag assignment vector jointly. Given the image $I$ we use a sliding window to split the image to $n_I$ over-lapping small image patches, $I\rightarrow \left [ \bfx_1,\cdots,\bfx_{n_I}\right ]$, where $\bfx_i$ is the visual feature vector of the $i$-th image patch. The convolutional representation of $I$ is given as
$
\bfy = \max(G)= [y_1,\cdots,y_r]^\top, where ~G = g\left ( W^\top X \right )
$,
where $W = [\bfw_1,\cdots,\bfw_r]$ is the matrix of $r$ filters, $g(\cdot)$ is a element-wise non-linear transformation function, and $\max(\cdot)$ gives the row-wise maximum elements, and $y_k$ is the maximum entity of the $k$-th row of $G$, $y_k = \max(G_{k,:})$. To learn the tag assignment vector $\bft$ of an image from its convolutional representation vector $\bfy$, we use a linear function to predict $\bft$ from $\bfy$,
$
\bft \leftarrow f(\bfy) =U\bfy - \bfb$, where $U\in \mathbb{R}$ and $\bfb$ are the parameters of the predictive model for the assignment vector. To the learn the CNN parameter $W$, linear predictor parameter $U$ and $\bfb$, and the complete tag matrix $T$, we consider propose the following minimization problem,
\begin{equation}
\label{equ:objec}
\begin{aligned}
&\min_{T,U,\bfb,W} \left\{ O(T,U,\bfb,W)= \sum_{i=1}^n Tr\left((\bft_i-\widehat{\bft}_i)^\top diag(\bfphi_i)(\bft_i-\widehat{\bft}_i)\right)
\vphantom{\sum_{i,i'=1}^n }
\right.\\
&\left.
+\lambda_1\sum_{i=1}^n \left\|\bft_i - (U\bfy_i - \bfb)\right\|_F^2
+\lambda_2 \sum_{i,i'=1}^n S_{ii'}\left\|\bft_i - \bft_{i'}\right\|_F^2
+\lambda_3 \sum_{i=1}^n \|\bft_i\|_1
 \right\}.
\end{aligned}
\end{equation}
The objective function terms are introduced as follows.
\begin{itemize}
\item The first term of the objective is to encourage the consistency between the available tags of $\widehat{\bft}_i$ and the estimated tag vector $\bft_i$ of an image $I_i$. It is defined as the The squared Frobenius norm distance between $t_{ji}$ and $\widehat{t}_{ji}$ weighted by the $\phi_{ji}$ is minimized with regard to $t_{ji}$. This term is popular in other tag completion works.

\item The second term is the squared Frobenius norm distance to measure the prediction error of the linear model of linear predictor. This loss term is novel and it has not been used in other works.

\item The third term is the visual similarity regularization term. For a images $I_i$, we seek its $k$-nearest neighbor set to present its visually similar images, denoted as $\mathcal{N}_i$. To measure the the similarity between $I_i$ and a neighboring image $I_{i'}\in \mathcal{N}_i$ is measured by the normalized Gaussian kernel of the Frobenius norm distance between their convolutional representation vectors,
$
S_{ii'} = \exp\left(-\gamma\|\bfy_i-\bfy_{i'}\|_F^2\right)/ \sum_{i''\in \mathcal{N}_i} \exp\left(-\gamma\|\bfy_i-\bfy_{i''}\|_F^2\right)$, if $I_{i'}\in~\mathcal{N}_i$, and $0$ otherwise.
If $S_{ii'}$ is large, the $I_i$ and $I_{i'}$ are visually similar, we expect their complete tag assignment vectors to be close to each other, and minimize the squared Frobenius norm distance between $\bft_i$ and $\bft_{i'}$ weighted by $S_{ii'}$. This term is not used by other tag completion works and it is novel in our work.

\item The last term is a sparsity term of the learned tag vectors, and it is also imposed by other works to seek the sparsity.

\end{itemize}
$\lambda_1$, $\lambda_2$, and $\lambda_3$ are the weights of different regularization terms of the objective.

To solve the problem in (\ref{equ:objec}), we use the gradient descent method and the alternate optimization strategy. In iterative algorithm, we first fix the variables to calculate the similarity measure $S_{ii'}$, and then fix the similarity measures to calculate the sub-/gradient regarding to different variables. The sub-/gradient functions with regard to the variables are calculated as

\begin{equation}
\label{equ:gradient}
\begin{aligned}
&\nabla_{\bft_i} O(\bft_i)= 2 diag(\bfphi_i)(\bft_i-\widehat{\bft}_i)
+2\lambda_1\left(\bft_i - (U\bfy_i - \bfb)\right)
+2\lambda_2 \sum_{i'=1}^n S_{ii'}\left(\bft_i - \bft_{i'}\right)\\
&
+2\lambda_3 \sum_{i=1}^n diag\left(\frac{1}{\left|t_{1i}\right|},\cdots,\frac{1}{\left|t_{mi}\right|}\right)\bft_i,\\
&
\nabla_{U}O(U)= - 2\lambda_1\sum_{i=1}^n \left(\bft_i - U\bfy_i + \bfb \right) \bfy_i^\top,
~\nabla_{\bfb}O(\bfb) = 2\lambda_1\sum_{i=1}^n \left(\bft_i - U\bfy_i + \bfb \right),\\
&
\nabla_{\bfw_k}O(\bfw_k) = \sum_{i=1}^n \left [ \nabla_{\bfy_i} O(\bfy_i)\right ]_k \nabla_{\bfw_k}
{\bfy_i(\bfw_k)}, \nabla_{\bfy_i} O(\bfy_i) = -2\lambda_1 U^\top \left(\bft_i - (U\bfy_i - \bfb)\right),\\
&
\nabla_{\bfw_k}\bfy_i(\bfw_k)=\nabla_{\bfw_k} g(\bfw_k^\top\bfx_{ij^*})\bfx_{ij^*},
~where~\bfx_{ij^*}={\arg\max}_{\bfx_i\in X_i}\bfw^\top \bfx_{ij}.
\end{aligned}
\end{equation}
For a variable, $x \in \{T,U,\bfb,W\}$, the updating rule is $x\leftarrow x - \eta \nabla O_x(x)$.

\section{Experiments}
\label{sec:exp}


In the experiments, we use three benchmark data sets of image, including Corel data set, Labelme data set, Flickr data set. The Corel data set has 4,993 images tagged by 260 unique tags, Labelme data set is composed of 2,900 of 495 tags, while Flickr data set has 1 million images of over 1,000 tags. We perform two groups of experiments, one group of image retrieval, and another group of image annotation.

\begin{table}[!htb]
\centering
\caption{Results of image annotation measured by Precision@5.}\label{tab:annot}
\begin{tabular}{|l||l|l||l|}
\hline
Data sets & Corel & Labelme & Flickr \\\hline\hline
Proposed & 0.47 & 0.30 &0.28 \\\hline
Lin et al. \cite{Lin20131618} & 0.35 & 0.22 & 0.18 \\\hline
Wu et al. \cite{Wu2013716} & 0.37 & 0.23 & 0.19 \\\hline
Feng et al.  \cite{Feng2014424} & 0.40 & 0.24 & 0.20 \\\hline
Lin et al. \cite{Lin201442} & 0.43 & 0.23 & 0.23 \\\hline
Li et al. \cite{Li2016425}& 0.44 & 0.25 & 0.24 \\\hline
\end{tabular}
\end{table}


\textbf{Image Annotation} Given an image, and a set of candidate tags, the problem of image annotation is to predict its true complete list of tags relevant to the image. This is an special case of image tag completion. We use the four-fold cross-validation protocol to split the training/ test subsets. We rank the tags for each image according to the tag scores output of our model, and the top-ranked tags are returned as the tags of the candidate image. The performance measures of Precision@5 is used to evaluate the results. We compare our method to the existing stat-of-the-art methods, including the methods proposed by Lin et al. \cite{Lin20131618}, Wu et al. \cite{Wu2013716}, Feng et al.  \cite{Feng2014424}, Lin et al. \cite{Lin201442}, and Li et al. \cite{Li2016425}. The results are reported in Table \ref{tab:annot}. From the results reported in Table \ref{tab:annot}, the proposed method outperforms the compared methods over all the three data sets on the four performance measures. This is an strong evidence of the advantage of the CNN model for the tag completion and annotation of images. This is not surprising because we use an effective convolutional reforestation method to extract features from the images, and the CNN parameters are tuned especially for the tag completion problem.

\begin{table}[!htb]
\centering
\caption{Results of image retrieval experiments measured by Pos@Top.}\label{tab:retreival}
\begin{tabular}{|l||c|c|c|}
  \hline
Methods & Corel &Labelme&Flickr \\\hline\hline
Proposed & 0.73 & 0.67 & 0.66 \\\hline
Lin et al. \cite{Lin20131618} & 0.64 & 0.57 & 0.54  \\\hline
Wu et al. \cite{Wu2013716} & 0.65 & 0.58 & 0.55  \\\hline
Feng et al.  \cite{Feng2014424} & 0.65 & 0.59 & 0.57 \\\hline
Lin et al. \cite{Lin201442} & 0.61 & 0.59 & 0.58 \\\hline
Li et al. \cite{Li2016425}& 0.68 & 0.62 & 0.60 \\\hline
\end{tabular}
\end{table}

\textbf{Image Retrieval}
Then we evaluate the proposed method over the problem of tag-based image retrieval \cite{liang2016optimizing}. This problem uses tags as queries to retrieve the images from the database of images. In each data set of images, we remove some tags of the images to set up the image tag completion problem, and then apply the image tag completion algorithm to complete the missing tags. We measure the retrieval performance by the positive at top (Pos@Top). The usage of this performance measure is motivated by the works of Liang et al. \cite{liang2016optimizing,li2016nuclear}. The works of Liang et al. \cite{liang2016optimizing,li2016nuclear} show that the Pos@Top is a robust and parameter-free performance measure, which is suitable for most database retrieval problems. Following the works of Liang et al. \cite{liang2016optimizing,li2016nuclear}, we adapt this performance measure to evaluate the results of the image retrieval problem in our experiments. The retrieval results of different methods are reported in Table \ref{tab:retreival}. We can observe from this table that the proposed method outperforms the other methods over all the three data sets.

\begin{figure}
  \centering
  \includegraphics[width=0.5\textwidth]{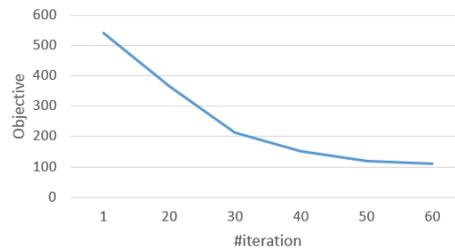}\\
  \caption{Convergence curve over Corel data set.}
  \label{fig:covergence}
\end{figure}

\textbf{Convergence of the alternating gradient descent} We also plot the curve of the objective values with regard to increasing iterations for the alternating gradient descent algorithm. The curve of experiments over the Corel data set is shown in Fig. \ref{fig:covergence}. According to Fig. \ref{fig:covergence}, the algorithm converge after 40 iterations.

\section{Conclusion and future works}
\label{sec:con}

In this paper, we proposed a novel image tag completion method. This method is based on the CNN model. We use the CNN model to represent the image, and then predict the complete image tags from the CNN representations. The complete tag assignment score vectors are also regularized by the visual similarities calculated from the CNN representations. We develop an iterative algorithm to learn the parameters of the CNN model, the linear predictive model, and the complete tags. The experiments of the problems of image annotation and image retrieval based on image tag completion over three benchmark data sets show the advantage of the proposed method. In the future, we will extend our work of CNN model to other machine learning problems beside image tag completion, such as computer vision \cite{liang2016novel,fan2016stochastic,tan2016robust,zhao2012connectivity,Mao2010,zhang2009detecting,zhang2016nonlinear}, material engineering \cite{wang2016donor,wolf2016benzo}, portfolio choices \cite{shen2017subset,shen2016portfolio,shen2015portfolio,shen2015transaction,shen2014doubly}, and biomedical engineering \cite{cai2011optimization,cai20162,cai2016low,cai2016class,Mao2008,li2015outlier,king2015surgical,mo2015importance,li2015burn,thatcher2016multispectral,shi2017nonlinear,hobbs2016quad}.

\end{document}